# THE "WHITEBOARD" ARCHITECTURE:
## A WAY TO INTEGRATE HETEROGENEOUS COMPONENTS OF NLP SYSTEMS

Christian Boitet

GETA, IMAG (UJF & CNRS),
150 rue de la Chimie, BP 53
38041 Grenoble Cedex 9, France
Christian.Boitet@imag.fr

Mark Seligman

ATR Interpreting Telecommunications Research Labs
2–2 Hikari-dai, Seika-cho, Soraku-gun
Kyoto 619-02, Japan
seligman@itl.atr.co.jp

**ABSTRACT**

We present a new software architecture for NLP systems made of heterogeneous components, and demonstrate an architectural prototype we have built at ATR in the context of Speech Translation.

## INTRODUCTION

Speech translation systems must integrate components handling speech recognition, machine translation and speech synthesis. Speech recognition often uses special hardware. More components may be added in the future, for task understanding, multimodal interaction, etc. In more traditional NLP systems, such as MT systems for written texts, there is also a trend towards distributing various tasks on various machines.

Sequential architectures [10, 11] offer an easy solution, but lead to loss of information and lack of robustness. On the other hand, reports on experiments with blackboard architectures [6, 13, 20] show they also have problems.

We are exploring an intermediate architecture, in which components are integrated under a *coordinator,* may be written in various programming languages, may use their own data structures and algorithms, and may run in parallel on different machines. The coordinator maintains in a *whiteboard* an image of the input and output data structures of each component, at a suitable level of detail. The whiteboard fosters reuse of partial results and avoids wasteful recomputation. Each component process is encapsulated in a *manager*, which transforms it into a *server*, communicating with external clients (including the coordinator) via a system of mailboxes. Managers handle the conversions between internal (server) and external (client) data formats. This protocol enhances modularity and clarity, because one needs to to explicitly and completely declare the appearance of the partial results of the components on the whiteboard.

Managers may also make batch components appear as incremental components by delivering outputs in a piecewise fashion, thus taking a first step towards systems simulating simultaneous translation.

We have produced a rudimentary architectural prototype, KASUGA, to demonstrate the above ideas.

In the first section, our four main guidelines are detailed: (1) record overall progress of components in a whiteboard; (2) let a coordinator schedule the work of components; (3) encapsulate components in managers; and (4) use the managers to simulate Incremental Processing. In the second section, some high-level aspects of the KASUGA prototype are first described, and a simple demonstration is discussed, in which incremental speech translation is simulated. Lower-level details are then given on some internal aspects.

## I. THE WHITEBOARD ARCHITECTURE

### 1. Record overall progress in a whiteboard

The whiteboard architecture is inspired by the chart architecture of the MIND system [8] and later systems or formalisms for NLP [1, 5], as well as by the blackboard architecture, first introduced in HEARSAY-II [6, 13] for speech recognition. However, there is a significant difference: the components do not access the whiteboard, and need not even know of its existence.

There are 2 main problems with the sequential approach.

- P1: loss of information

  If components are simply concatenated, as in Asura [10, 11], it is difficult for them to share partial results. Information is lost at subsystem interfaces and work has to be duplicated. For example, the cited system uses an LR parser to drive speech recognition; but syntactic structures found are discarded when recognition candidates are passed to MT. Complete reparsing is thus needed.

- P2: lack of robustness

  Communication difficulties between subsystems may also damage robustness. During reparsing for MT in ASURA, if no well-formed sentences are found, partial syntactic structures are discarded before semantic analysis; thus there is no chance to translate partially, or to use semantic information to complete the parse.

The pure blackboard approach solves P1, but not P2, and introduces four other problems.

- P3: control of concurrent access

  In principle, all components are allowed to access the blackboard: complex protection and synchronization mechanisms must be included, and fast components may be considerably slowed down by having to wait for permission to read or write.

- P4: communication overloads

  The amount of information exchanged may be large. If components run on different machines, such as is often the case for speech-related components, and may be the case for Example-Based MT components in the future, communication overloads may annihilate the benefit of using specialized or distributed hardware.

- P5: efficiency problems

  As components compute directly on the blackboard, it is a compromise by necessity, and can not offer the optimal kind of data structure for each component.

- P6: debugging problems

  These are due to the complexity of writing each component with the complete blackboard in mind, and to the parallel nature of the whole computation.

In the "whiteboard" approach, the global data structure is hidden from the components, and accessed only by a "coordinator". (The whiteboard drawing is expanded later.)

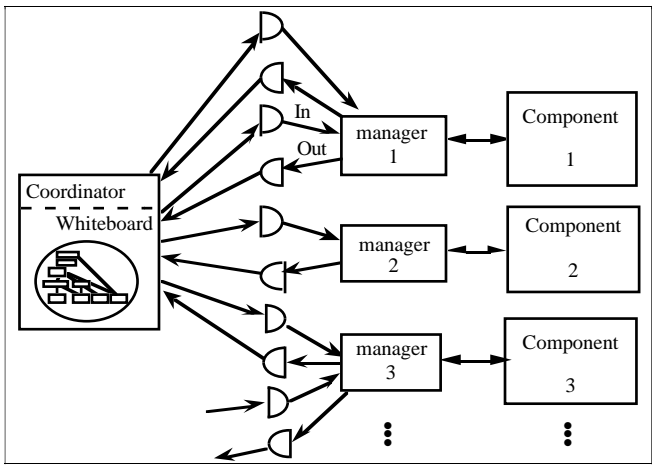

*Figure 1: the "whiteboard" architecture*

This simple change makes it possible to avoid problems P3–P6. It has also at least two good points:
- *It encourages developers to clearly define and publish what their inputs and outputs are,* at least to the level of detail necessary to represent them in the whiteboard.
- *The whiteboard can be the central place where graphical interfaces are developed* to allow for easy inspection, at various levels of detail.

As long as an NLP system uses a central record accessed only by a "coordinator" and hidden from the "components", it can be said to use a whiteboard architecture. It remains open what data structures the whiteboard itself should use.

As in [2], we suggest the use of a time-aligned lattice, in which several types of nodes can be distinguished. In stating our preference for lattices, we must first distinguish them from grids, and then distinguish true lattices from 2 types of quasi-lattice, charts and Q-graphs (fig. 2 & 3).

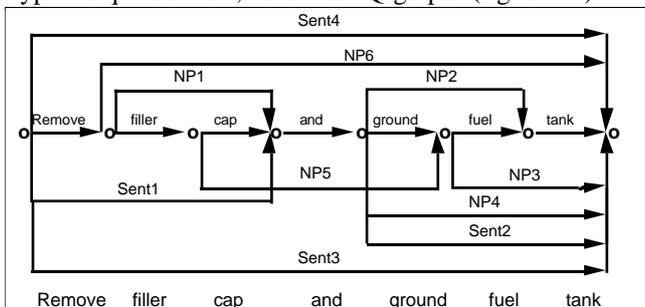

*Fig. 2: chart built on a syntactically ambiguous sentence*

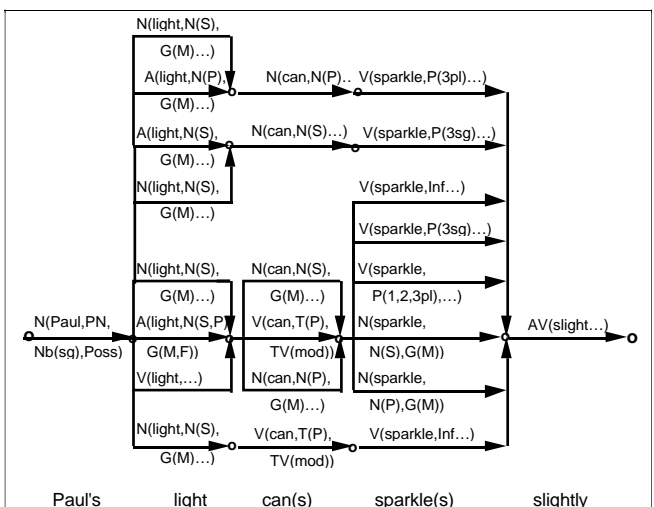

*Fig. 3: A Q-graph for a phonetically ambiguous sentence*

*Grids* have no arcs, but nodes corresponding to time spans. A node N spanning [t1,t2] is implicitly connected to another node N' spanning [t'1,t'2] iff its time span begins earlier (t1≤t'1), ends strictly earlier (t2<t'2), and the respective spans (a) are not too far apart and (b) don't overlap too much (t2–max-gap≤t'1≤t2+max-overlap). max-gap and max-overlap are *gapping* and *overlapping thresholds* [12]. Because t2<t'2, there can be no cycles.

In a *lattice,* by contrast, nodes and arcs are explicit. Cycles are also forbidden, and there must be a unique first node and a unique last node.

Grids have often been used in NLP. For example, the output of the phonetic component of KÉAL [12] was a word grid, and certain speech recognition programs at ATR produce phoneme grids[1]. In general, each node bears a time span, a label, and a score. Grids can also be used to represent an input text obtained by scanning a bad original, or a stenotypy tape [9], and to implement some working structures (like that of the Cocke algorithm).

However, we will require explicit arcs in order to explicitly model possible *sequences*, sometimes with associated information concerning sequence probability. Thus raw grids are insufficient for our whiteboards.

Two kinds of quasi-lattices have been used extensively, in two varieties. First, *chart structures* have originally been introduced by M. Kay in the MIND system around 1965 [8]. In a chart, as understood today (Kay's charts were more general), the nodes are arranged in a row, so that there is always a path between any two given nodes. The arcs bear the information (label, score), not the nodes. Charts are also used by many unification-based natural language analyzers [14].

Chart structures are unsuitable for representing results on a whiteboard, however, because they are unable to represent *alternate sequences*. Consider the alternate word sequences of Figure 4. It is not possible to arrange the words in a single row so that all and only the proper sequences can be read out.

| I would like | it if you you to | came come be early | early earlier | tomorrow |
|---|---|---|---|---|

*Figure 4: A sentence with alternate formulations*

A second type of quasi-lattice is the *Q-graphs* of [5] and their extension [17], the basic data structure for text representation in the METEO [4] and TAUM-Aviation [7] systems. A Q-graph is a loop-free graph with a unique entry node and a unique exit node. As in charts, the information is carried on the arcs. It consists in labeled or annotated trees. As there may be no path between two nodes, Q-graphs can indeed faithfully represent alternate sequences like those of Figure 4. But in this case it is necessary to use, on more than one arc, identical labels referring to the same span of the input. For representation on a whiteboard, such duplication is a drawback.

To simplify bookkeeping and visual presentation, we prefer a representation in which *a given label referring to a given span appears in only one place.* A true lattice, like that of Figure 5, makes this possible.

The decomposition of the lattice in layers seems natural, and leads to more clarity. Each layer contains results of one component, selected to the "appropriate level of detail". Its time-aligned character makes it possible to organize it in such a way that everything which has been computed on a certain time interval at a certain layer may be found in the same region. Each layer has three dimensions, *time, depth* and *label* (or "class"). A node at position (i,j,k) corresponds to the input segment of length j ending at time i and is of label k. All realizations of label k corresponding to this segment are to be packed in this node, and all nodes corresponding to approximately equal input segments are thus geometrically clustered.

---

[1] [15, 16]. By contrast, the HWIM [20] system used a "phonetic lattice" on which an extended ATN operated.

In other words, ambiguities are packed so that *dynamic programming techniques may be applied on direct images of the whiteboard.* Figure 6 gives an example, where the main NP has been obtained in two ways.

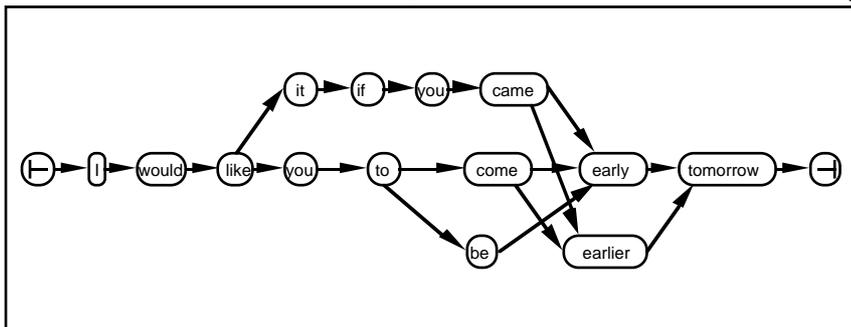

*Figure 5: A word lattice (representing a sentence with alternate formulations)*

Arcs may optionally be augmented with activation or inhibition weights, so that ideas from the fast-developing field of neural networks may be applied.

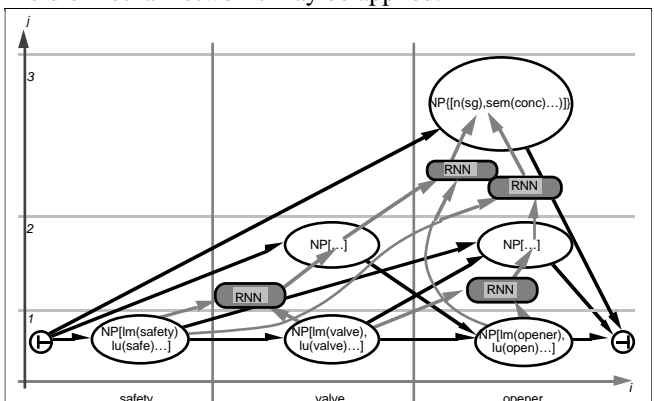

*Figure 6: The whiteboard as a factorizing data structure*

The true lattice, then, is our preferred structure for the whiteboard.

We said that the whiteboard could be a central place for transparent inspection, at suitable levels of detail. We use the notion of "shaded nodes" for this.

- "White" nodes are the real nodes of the lattice. They contain *results* of the computation of the component associated with their layer: a white node contains at least a label, legal in its layer, such as NP, AP, CARDP, VP… in the example above, and possibly more complex information, as allowed by the declaration of the layer in the whiteboard.
- "Grey" nodes may be added to show how the white nodes have been constructed. They don't belong to the lattice structure proper. In the example above, they stand for *rule instances,* with the possibility of m—>n rules. In other cases, they may be used to show the correspondences between nodes ot two layers.

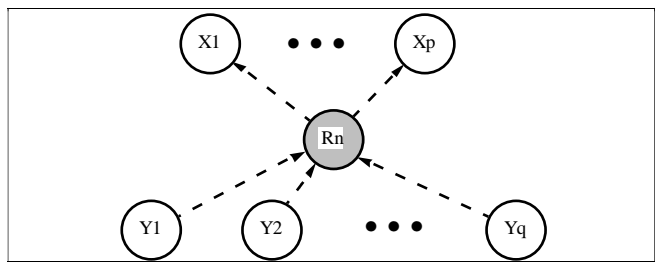

*Figure 7: White and grey nodes corresponding to rule Rn: X1 X2…Xp –> Y1 Y2…Yq*

- "Black" nodes may be used to represent finer steps in the computation of the component, e.g. to reflect the active edges of a chart parser.

Whiteboard layers are organized in a loop-free dependency graph. Non-linguistic as well as linguistic information can be recorded in appropriate layers. For example, in a multimodal context, the syntactic analyzer might use selected information from a map layer, where pointing, etc. could be recorded. Interlayer dependencies should be declared, with associated constraints, stating for instance that only nodes with certain labels can be related to other layers. Here is an illustration of that idea, without any pretense to propose a realistic choice of layers, however.

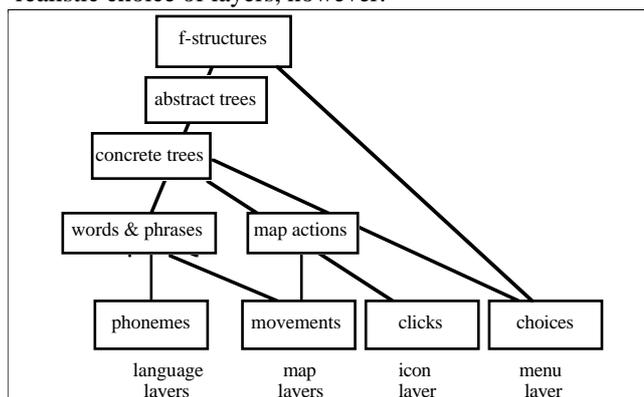

*Figure 8: A hierarchy of layers in an hypothetical whiteboard for a multimodal NLP system*

## 2. Let a coordinator schedule the components

In its simplest form, a coordinator only transmits the results of a component to the next component(s). However, it is in a position to carry out global strategies by filtering low-ranking hypotheses and transmitting only the most promising part of a whiteboard layer to its processing component. Further, if certain components make useful predictions, the coordinator can pass these to other components as constraints, along with input.

## 3. Encapsulate components in managers

Developers of components should be free to choose and vary their algorithms, data structures, programming languages, and possibly hardware (especially so for speech-related components). Our approach is to encapsulate existing components in *managers*, which hide them and transform them into servers. This strategy has the further advantage of avoiding any direct call between coordinator and components. To plug in a new component, one just writes a new manager, a good part of which is generic.

A manager has a request box where clients send requests to open or close connections. A connection consists of a pair of in and out mailboxes, with associated locks, and is opened with certain parameters, such as its sleep time and codes indicating pre-agreed import and export formats. The coordinator puts work to do into in-boxes and gets results in corresponding out-boxes.

As illustrated in Figure 1 above, a client can open more than one connection with the same manager. For example, an on-line dictionary might be called for displaying "progressive" word for word translation, as well as for answering terminological requests by a human interpreter supervising several dialogues and taking over if needed. And a manager can in principle have several clients. However, this potential is not used in KASUGA.

### 4. Simulate incremental processing

In real life, simultaneous interpretation is often preferred over consecutive interpretation: although it may be less exact, one is not forced to wait, and one can react even before the end of the speaker's utterance. Incremental processing will thus be an important aspect of future machine interpretation systems. For instance, a semantic processor might begin working on the syntactic structures hypothesized for early parts of an utterance while later parts are still being syntactically analyzed [19].

Even if a component (e.g., any currently existing speech recognizer) has to get to the end of the utterance before producing any result, its manager may still make its processing appear incremental, by delivering its result piecewise and in the desired order. Hence, this organization makes it possible to simulate future incremental components.

## II. THE KASUGA PROTOTYPE

### 1. External level

The coordinator (KAS.COORD) is written in KEE™, an object-oriented expert system shell with excellent interface-building tools. The whiteboard is declared in KEE's object language. KEE itself is written in Common Lisp.

Three components are involved:
- speech recognition (SP.REC) providing a 3-level grid, programmed in C [15];
- island-driven syntactic chart-parsing (SYNT.AN) deriving words and higher-level syntactic units, programmed in C;
- word-for-word translation (WW.TRANS) at the word level, written in C and running on another machine.

The managers are written in Lisp, and run independently, in three Unix processes. Each manager and the coordinator can run in different Unix shells. Although WW.TRANS is already accessible as a server on a distant machine, we had to create a manager for it to get the intended behavior.

With only these components, it is possible to produce a simple demonstration in which incremental speech translation is simulated and the transparency gained by using a whiteboard is illustrated. The phonemes produced by SP.REC are assembled into words and phrases by SYNT.AN. As this goes on, WW.TRANS produces possible word-for-word translations, which are presented on screen as a word lattice.

KASUGA's whiteboard has only three layers: phonemes; source words and phrases; and equivalent target words. At the first layer, the phoneme lattice is represented with phonemes in nodes. At the second layer, we retain only the complete substructures produced by SYNT.AN, that is, the inactive edges. Phonemes used in these structures appear again at that layer.

In KEE, we define a class of NODES, with subclasses WHITE.NODES, GREY.NODES, PHON.LAYER.NODES, and SYNT.LAYER.NODES in the syntactic layer. NODES have a generic display method, and subclasses have specialized variants (e.g., the placing of white nodes depends on their time interval, while that of grey nodes depends on that of the white nodes they connect).

### 2. Internal level

When a manager receives a Make.Connection request from a client, it creates an in box and an out box (and associated locks, used to prevent interference between components), through which information is passed to and from the client. The Make.Connection request includes codes showing in which format(s) the client is expecting to deposit data in the in box and read data from the out box, for that connection.

Although data transfer could be programmed more efficiently, e.g. using Unix sockets, our method is more general, as it uses only the file system, and we believe its overhead will be negligible in comparison with the processing times required by the components.

For each out box, the client (KASUGA) activates a reader process and the relevant manager activates a writer process. Conversely, for each in box, the client activates a writer process and the manager activates a reader process. A reader process wakes up regularly and checks whether its mailbox is both non-empty and unlocked. If so, it locks the mailbox; reads its contents; empties the mailbox; unlocks it; and goes to sleep again. A writer process, by comparison, wakes up regularly and checks whether its mailbox is both empty and unlocked. If so, it locks the box, fills it with appropriate data, unlocks it, and goes back to sleep. For example, the writer associated with SYNT.AN will deposit in the appropriate out box the image of all the inactive arcs created since the last deposit.

SP.REC provides, for each of 40 prerecorded bunsetsu (elementary phrase), a set of about 25 phoneme matrices, one for each phoneme. A matrix cell contains the score for a given phoneme with a given beginning/ending speech frame pair. These matrices are then compared, and 3 other matrices are computed. The top-scoring matrix contains in each cell the top-scoring phone and its score for the corresponding beginning/end. The 2nd-scoring and 3rd-scoring matrices are computed similarly. These three matrices are used to build the first layer of the whiteboard.

To build the whiteboard's second layer, an island-driven chart parser is used, where the matrices are considered as initialized charts. The overall best-scoring cell in the top matrix is established as the only anchor, and bi-directional searching is carried out within the (handset) limits set by max-gap and max-overlap. A CFG written by J. Hosaka for the ASURA demos is now used as is. Parsing results are converted to syntactic.lattice.N (by our chart-to-lattice filter) and brought into KEE.

Then an image lattice, WW.lattice.N, is computed as the whiteboard's third layer, using a C-based on-line J-E dictionary. Each lexical syntactic node gives rise to one English word for each meaning. For example, <u>hai</u> gives yes, yes-sir, the-lungs, ashes, etc.

Layers of the whiteboard are represented by KEE "planes". We can move planes relative to each other; zoom in various ways; put various information in the nodes (label, rule responsible, id, time span, score); expand the nodes; open & close the nodes selectively. And we can color the nodes according to their score. It is possible to show or hide various parts of the whiteboard. In Figure 9, the first layer, the time grid, the lattice lines, and the initial/final lattice nodes have been hidden. Alternatively, we could hide construction (dotted) lines, rule boxes, label boxes, etc. The view of any part of the whiteboard can be changed for emphasis: one can for instance interactively select only the nodes above a certain confidence threshold. Overall processing can be interrupted for examination.

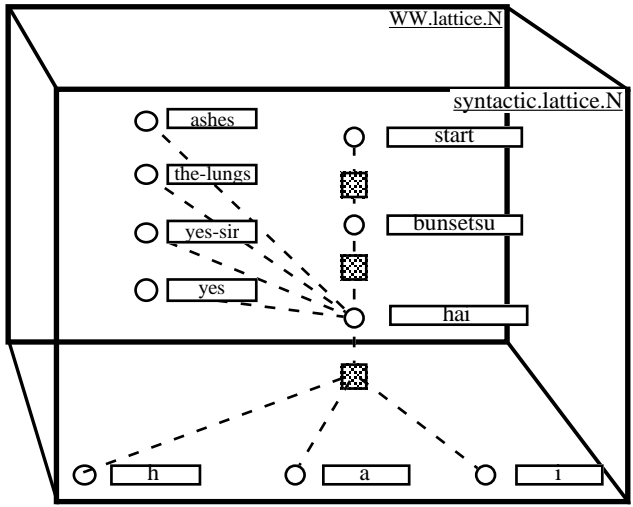

*Figure 9: a view of KASUGA's whiteboard*

If this architecture is to be further developed in the future, one could use instead of KEE a general-purpose, portable interface building toolkit in order to avoid the overhead and overspecialization associated with using a complete expert system shell.

KAS.COORD writes and reads data to and from the managers in a LISP-like format, and handles the transformation into KEE's internal format. Each manager translates back and forth between that format and whatever format its associated component happens to be using. Hence, formats must be precisely defined. For instance, the edges produced by the speech recognizer are of the form (begin end phoneme score). The nodes and edges of the corresponding phoneme layer in the whiteboard are of the form (node-id begin end phoneme score (in-arcs) (out-arcs)), with arcs being of the form (arc-id origin extremity weight).

## CONCLUSION

Although the concept of the whiteboard architecture has emerged in the context of research in Speech Translation, it can be useful in other areas of NLP. It has already been used, in a preliminary form, in dialogue-based MT [3]: the tasks are distributed between the authoring stations and an MT server, and the coordinator maintains in a unique data structure all intermediate stages of processing of all units of translation.

The whiteboard architecture might be used with profit in all situations where it is important to integrate new or existing components, e.g. to build generic environments for developing heterogeneous NLP systems. Researchers would thereby gain twice: by getting a clearer view of what they (and others) are doing; and by being able to use generic interface tools provided by the coordinator for debugging and illustrating purposes.

**ACKNOWLEDGMENTS**

We are grateful to M. Fiorentino from Intellicorp, Inc. and K. Kurokawa from CSK, Inc., for providing a demo copy of KEE™ and valuable technical support; to Dr. Y. Yamazaki, President of ATR-IT, and T. Morimoto, Head of Dept. 4, for their support and encouragement; to H. Singer, T. Hayashi, Y. Kitagawa, H. Kashioka, and J. Hosaka, for their help in developing the components; and to K. H. Loken-Kim, for stimulating discussions and proposing the term "whiteboard".

-o-o-o-o-o-o-o-o-o-o-